\PassOptionsToPackage{hyphens}{url}
\PassOptionsToPackage{shortlabels,inline}{enumitem}
\documentclass[10pt,twocolumn,letterpaper]{article}

\usepackage[pagenumbers]{cvpr} 

\usepackage[utf8]{inputenc}
\usepackage[font=small]{caption}
\usepackage{amsmath,amssymb,amsthm,nccmath,mathtools} 
\usepackage{booktabs,multirow,adjustbox}             
\usepackage{algorithm,algpseudocode}                 
\usepackage[switch]{lineno}
\usepackage{microtype}
\usepackage{graphicx}
\usepackage{academicons}
\usepackage{csquotes}
\usepackage[dvipsnames]{xcolor} 
\usepackage{lipsum}
\usepackage{subcaption}
\usepackage{dblfloatfix}
\usepackage{enumitem} 

\usepackage{graphicx}
\usepackage{amsmath,amssymb}
\usepackage{booktabs}
\usepackage{multirow}
\usepackage{float}
\usepackage{enumitem}
\usepackage[dvipsnames]{xcolor}

\usepackage{algorithm}
\usepackage{algpseudocode}
\usepackage{url}

\definecolor{cvprblue}{rgb}{0.21,0.49,0.74}
\usepackage[breaklinks,colorlinks,allcolors=cvprblue]{hyperref}
\def\paperID{*****} 
\def\confName{CVPR}
\def\confYear{2026}
\title{Contrast-Enhanced Gating in GRUs for Robust Low-Data Sequence Learning}
\author{
  Barathi Subramanian$^{1}$, Rathinaraja Jeyaraj$^{1}$, Anand Paul$^{2}$ \\
  $^{1}$Stanford University, USA, $^{2}$LSU Health Sciences Center New Orleans, USA \\
  \tt\small \{barathi1, rajaj\}@stanford.edu, apaul4@lsuhsc.edu
}
\vspace{-6pt}
\begin{document}

\maketitle 
\vspace{-5pt}
\begin{abstract}
Activation functions govern how recurrent networks regulate and transmit information across temporal dependencies. Despite advances in sequence modelling, gated recurrent units (GRUs) still depend on the standard sigmoid and tanh nonlinearities, which can produce weak gate separation and unstable learning, particularly when training data are limited. We introduce squared sigmoid-tanh (SST), a parameter-free activation that squares the gate nonlinearity to increase contrast between near-zero- and high-activations, thereby promoting sharper information filtering during GRU updates. We incorporate SST into GRU gating and evaluate it across low-data settings spanning sign language recognition, human activity recognition, and time-series forecasting and classification. Across tasks, SST-GRU consistently surpasses standard sigmoid/tanh GRU, with the largest improvements observed in the smallest-data domains, while adding negligible computational cost. We further examine gate activation statistics and training dynamics, showing that SST improves training stability, which aligns with its performance gains in data-scarce settings. SST is a parameter-free modification that complements more complex architectural advances by improving gating selectivity in low-data sequence learning.
\end{abstract}
\vspace{-10pt}

\section{Introduction}
Activation functions determine how neural networks transform intermediate signals and are central to representational capacity and optimization stability in deep learning \cite{ref1}, \cite{ref2}. In feedforward networks, piecewise-linear activation functions such as ReLU \cite{ref3} and its variants \cite{ref4}, \cite{ref5} are widely used due to their simple piecewise behaviour and favourable gradient propagation, enabling deep models to train reliably at scale. In contrast, recurrent neural networks (RNNs) must propagate information through time, where training is notoriously sensitive to vanishing and exploding gradients \cite{ref6}. Early analyses showed that gradient-based learning becomes increasingly difficult as temporal dependencies grow, producing brittle optimization when long-range credit assignment is required. Later work further formalized these failure modes and proposed practical stabilizers (for example, gradient clipping and regularization strategies) to mitigate exploding or vanishing gradients in recurrent model training \cite{ref7}. Gated architectures such as long short-term memory (LSTM) \cite{ref8} was introduced to address these issues by improving error flow through time, and remains foundational sequence models. To handle low-data availability and reduce computational complexity, gated recurrent unit (GRU)-style architectures\cite{ref9}, \cite{ref10} were introduced to learn sparse patterns in the dataset. GRU relies primarily on sigmoid and tanh nonlinearities \cite{ref11}, \cite{ref12} to learn temporal dependencies effectively. However, in low-data regimes, these saturating nonlinearities may reduce separability between \enquote{open} and \enquote{closed} gate states, limiting the model’s ability to sharply filter or preserve information, resulting in reduced downstream performance \cite{ref6}, \cite{ref13}. This limitation becomes particularly relevant in compact sequential models deployed in resource-constrained environments such as gesture recognition pipelines \cite{ref14} and sensor-based time-series modelling. Rather than increasing model capacity or introducing new architectural components, this work investigates whether improving gate selectivity alone can enhance sequence learning under limited data conditions. 

Modern transformer-family models \cite{ref15}, \cite{ref16} have become strong general-purpose sequence learners, but they introduce practical trade-offs that matter in the target applications. First, self-attention can be computationally expensive for long sequences \cite{ref9}, and many time-series variants focus on architectural changes to reduce parameterization and improve efficiency (e.g., factorized or lightweight designs), which implicitly acknowledges that vanilla transformers can be over-parameterized or inefficient for typical time-series inputs. Second, transformers \cite{ref16}, \cite{ref17} and hybrid alternatives (including state-space inspired models \cite{ref18}, \cite{ref21}) show that the field is actively searching for architectures that retain accuracy while reducing compute and improving robustness for time-series applications. Third, evidence is mixed on whether scaling or pretraining alone guarantees superior performance over smaller alternatives in time-series settings, motivating careful choices under limited data and constrained training budgets \cite{ref19}, \cite{ref27}. Lightweight encoder-only designs aim to retain accuracy while reducing compute for forecasting at scale. Broader evaluations also report mixed outcomes on whether larger or pretrained transformers consistently dominate smaller alternatives in time-series settings, motivating careful architecture choices when data and training data are limited. In addition, hybrid state-space models \cite{ref18} highlight the continued push toward models that remain robust and efficient on long sequences.

In applications such as sign language recognition \cite{ref14}, human activity recognition \cite{ref30}, \cite{ref31}, speech recognition \cite{ref22}, \cite{ref23}, and time-series classification/forecasting \cite{ref33}, where datasets are limited, GRU remains attractive because they are compact, stable to train, and fast to deploy. However, data complexity makes GRU underfit the sequential data, which requires enhancing the model’s ability to learn complex patterns. This motivates a complementary direction to improve the gating nonlinearity itself while keeping the GRU structure unchanged to improve the downstream task performance. To this end, we propose squared sigmoid-tanh (SST), a simple drop-in activation strategy that squares the outputs of sigmoid and tanh within the GRU gating pathway. Squaring increases the contrast between weak and strong activations (e.g., values near 0 are suppressed further while larger activations are emphasized), which can encourage sharper information filtering through the update dynamics to improve the model generalizability and downstream performance. Importantly, SST does not introduce additional parameters and adds negligible computational overhead, making it compatible with resource-constrained training and deployment. We empirically evaluate SST-powered GRU across multiple low-data tasks and show consistent gains over standard GRU nonlinearities. In summary, the following contributions are presented in this manuscript. We  

\begin{itemize}
    \item introduce, SST, a drop-in nonlinear modification for GRU gating designed for data-constrained sequence learning. 
    \item provide an analysis of how SST changes gate activation behaviour (e.g., gate contrast/distribution) and discuss practical implications for stability and convergence.
    \item benchmark SST-powered GRU across multiple applications (time-series analysis, sign language and human activity recognition) under controlled low-data protocols.
\end{itemize}

\begin{figure*}[t!]
\centering
    \includegraphics[width=.8\textwidth]{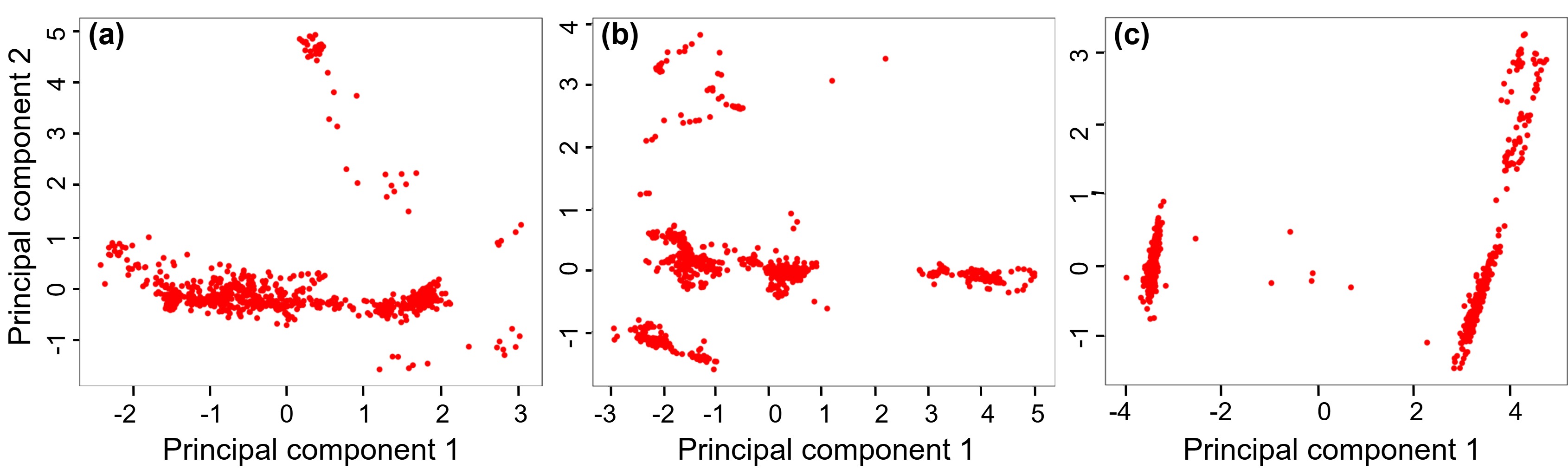} 
    \caption{PCA visualization of frames for three distinct ISL classes (a) friend, (b) phone call, and (c) location.}
    \vspace{-10pt}
    \label{fig:1}
 \end{figure*} 

\section{Related Work}
\textit{Activation functions and optimization}: Activation functions strongly influence gradient flow and training stability \cite{ref6}. Rectifier-based activations became prevalent in deep feedforward networks because they reduce the optimization difficulties caused by saturating nonlinearities, particularly vanishing gradients, while encouraging sparse activations and improved gradient flow \cite{ref19}. This line of work motivated numerous rectifier variants, including learnable negative-slope units (PReLU) \cite{ref5} and smooth alternatives such as ELU \cite{ref26}, which explicitly contrasts the contractive behaviour of tanh/sigmoid with rectifier-like units.  Beyond hand-designed functions, automated search has also produced drop-in nonlinearities such as Swish \cite{ref27}, which can outperform ReLU in several deep-network settings. These advances, however, largely target feedforward optimization and do not directly resolve the temporal credit-assignment challenges characteristic of recurrent learning. \\[2 pt]
\textit{Recurrent learning and gating under limited data}: Training RNNs on long-term dependencies remains difficult due to vanishing and exploding gradients \cite{ref12}, \cite{ref25}. Practical analyses further show that recurrent optimization can be brittle and often requires stabilizers such as careful initialization and gradient control \cite{ref2}. Gated models, notably LSTMs, were introduced to improve error flow through time and remain widely used for sequential data \cite{ref8}. Yet, common gated RNNs still use sigmoid and tanh to implement bounded gating and state updates, meaning saturation effects can still constrain gate contrast and gradient propagation that are amplified when supervision is scarce (small datasets) or sequences are short/noisy \cite{ref7}. \\[2 pt]
\textit{Learned and parameterized activations}: Adaptive piecewise linear (APL) units parameterize a flexible piecewise  nonlinearity \cite{ref28}. While such methods increase representational flexibility, they introduce additional learned parameters and typically validated in feedforward settings, leaving open the question of whether simple nonlinear modifications can improve recurrent gating in low-data settings. \\[2 pt]
\textit{Efficient recurrent cells and alternatives to standard RNNs}: Recent work also revisits the recurrent cell design to improve efficiency and parallelism. Quasi-RNNs combine convolutional computation with lightweight recurrent pooling to increase throughput while maintaining order sensitivity \cite{ref29}. Simple recurrent units \cite{ref20} target parallelizable recurrence and report substantial speedups over LSTMs in NLP tasks \cite{ref27} modifying the cell architecture. While alternative activations such as Swish and other smooth nonlinearities improve optimization in feedforward networks, recurrent gating still relies predominantly on sigmoid and tanh due to their bounded outputs. This highlights an opportunity to improve gate selectivity through simple contrast-enhancing transformations without increasing model complexity. \\[2 pt]
While there are attempts to introduce specialized recurrent activations, the existing techniques do not fully address the constraints posed by sequential data, particularly in sparse, low-data regimes. Therefore, we propose SST, explicitly designed to enhance training and generalization in classical GRU, when dealing with small sequential datasets containing sparse patterns. SST squares the outputs of the sigmoid and tanh activation functions, thereby augmenting gradient propagation and signal filtering, which is crucial for memory retention as signals evolve over time. This approach is directly targeted at mitigating the long-range dependency limitations inherent in existing activation schemes. Through our empirical studies, we demonstrate SST's effectiveness in enhancing classical GRU, providing new insights and empirical evidence for an open problem in neural networks. 
\section{Motivation and Intuition behind SST}
In certain cases, the performance of sequential models can be limited when training data exhibit sparse patterns. sigmoid and tanh nonlinearities, due to their saturation behaviour, may suppress gradient flow and reduce sensitivity to subtle variations in temporal signals. This is particularly relevant in tasks such as sign language recognition, where informative patterns may occur infrequently and require careful preservation during learning. To illustrate this challenge, principal component analysis (PCA) is applied to the Indian sign language (ISL) dataset \cite{ref14}, as shown in Figure \ref{fig:1}. Three representative sign classes (friend, phone call, location) are selected for visualization. The PCA projections show regions where feature representations are relatively sparse, indicating subtle temporal distinctions that conventional gating nonlinearities may fail to capture. To address this limitation, SST is introduced to improve sensitivity to weak but informative signals. Squaring the activation sharpens contrast between weak and strong gate responses, improving information selectivity in recurrent updates. While alternative sharpening strategies (e.g., $sigmoid(x)^p$) could also modify gate behaviour, the squared form is chosen as a minimal, smooth, and parameter-free transformation that preserves stable optimization. 
\section{Squared sigmoid-tanh (SST)}
\begin{figure}[b!]
    \vspace{-10pt}
    \centering
    \includegraphics[width=.4\textwidth]{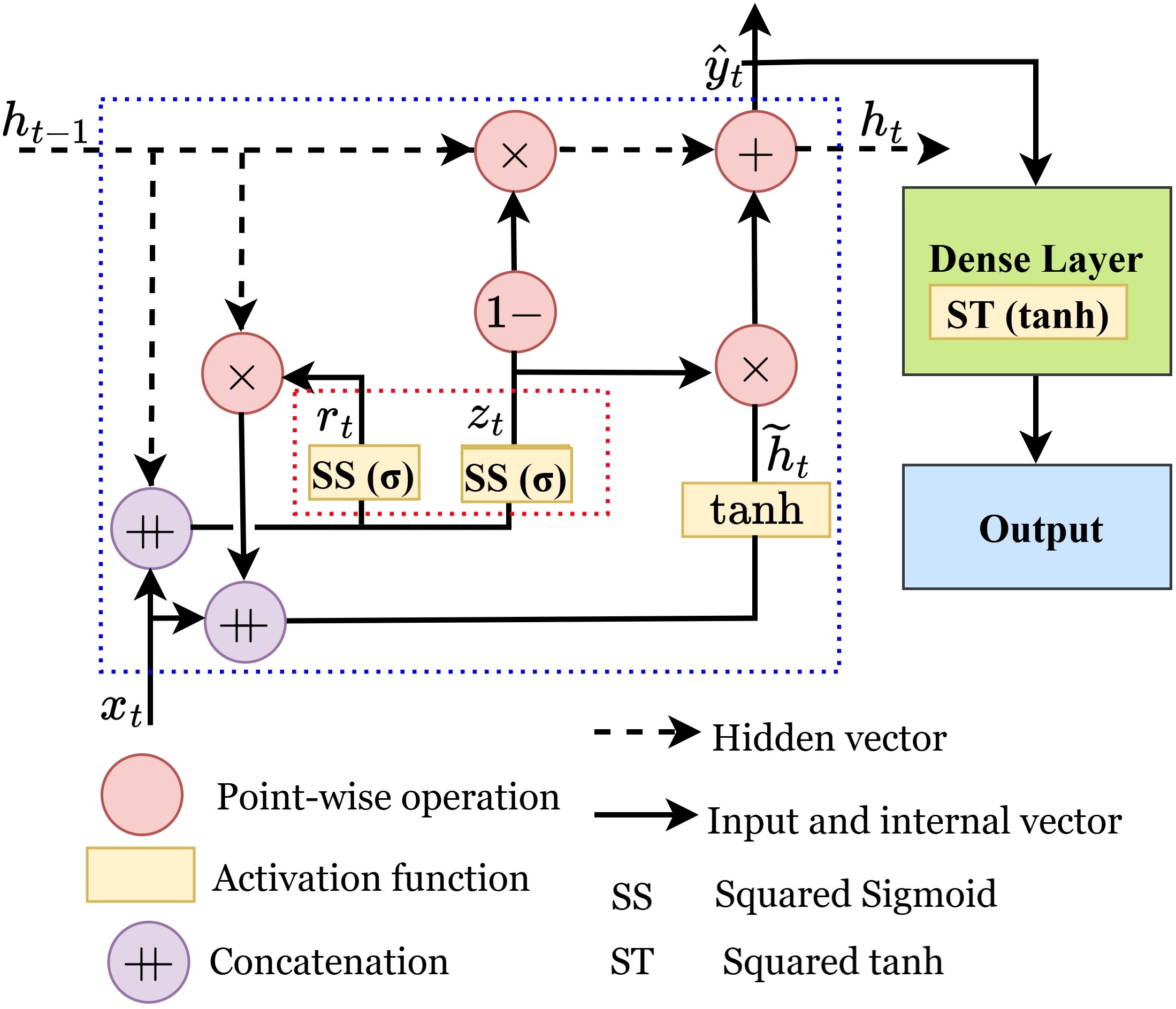}  
    \caption{SST-activation function in a GRU cell.}
    \label{fig:2}
 \end{figure}

The classical GRU plays a crucial role in modeling temporal dependencies, especially in modelling short-term and long-term sequences. A GRU cell contains two main components: a reset gate ($r_{t}$) and an update gate ($z_{t}$). These gates modulate the flow of information inside the GRU cell using sigmoid activation, controlling whether to forget the previous state or consider the latest input. The output ($h_{t}$) is a linear interpolation between the previous activation ($h_{t-1}$) and the GRU's candidate activation ($\widetilde{h}_{t}$) that processes the current input and previous state through a tanh layer. Although GRU exhibits a remarkable capacity to perform well in situations where training sequential data is limited \cite{ref7}, the classical GRU struggles to learn sparse patterns in the data and is unable to effectively propagate signals through time and layers \cite{ref27}. To address this issue, a novel idea, SST is introduced. SST squares the sigmoid activation in the update and reset gates, and the tanh activation in the final dense layer, while the candidate hidden state activation within the GRU remains unchanged, as shown in Figure \ref{fig:2}. The idea of SST is that, for instance, after squaring sigmoid activation function, the higher input probability value gets relatively higher than the lower input probability value. This approach enhances model learning efficiency and improves the accuracy of the neural network. In the following subsections, we discuss the properties of the squared sigmoid and squared tanh activation functions to formulate the SST.  

\subsection{Squared sigmoid (SS)}
\begin{figure}[b!]
 \vspace{-10pt}
    \centering
    \includegraphics[width=.4\textwidth]{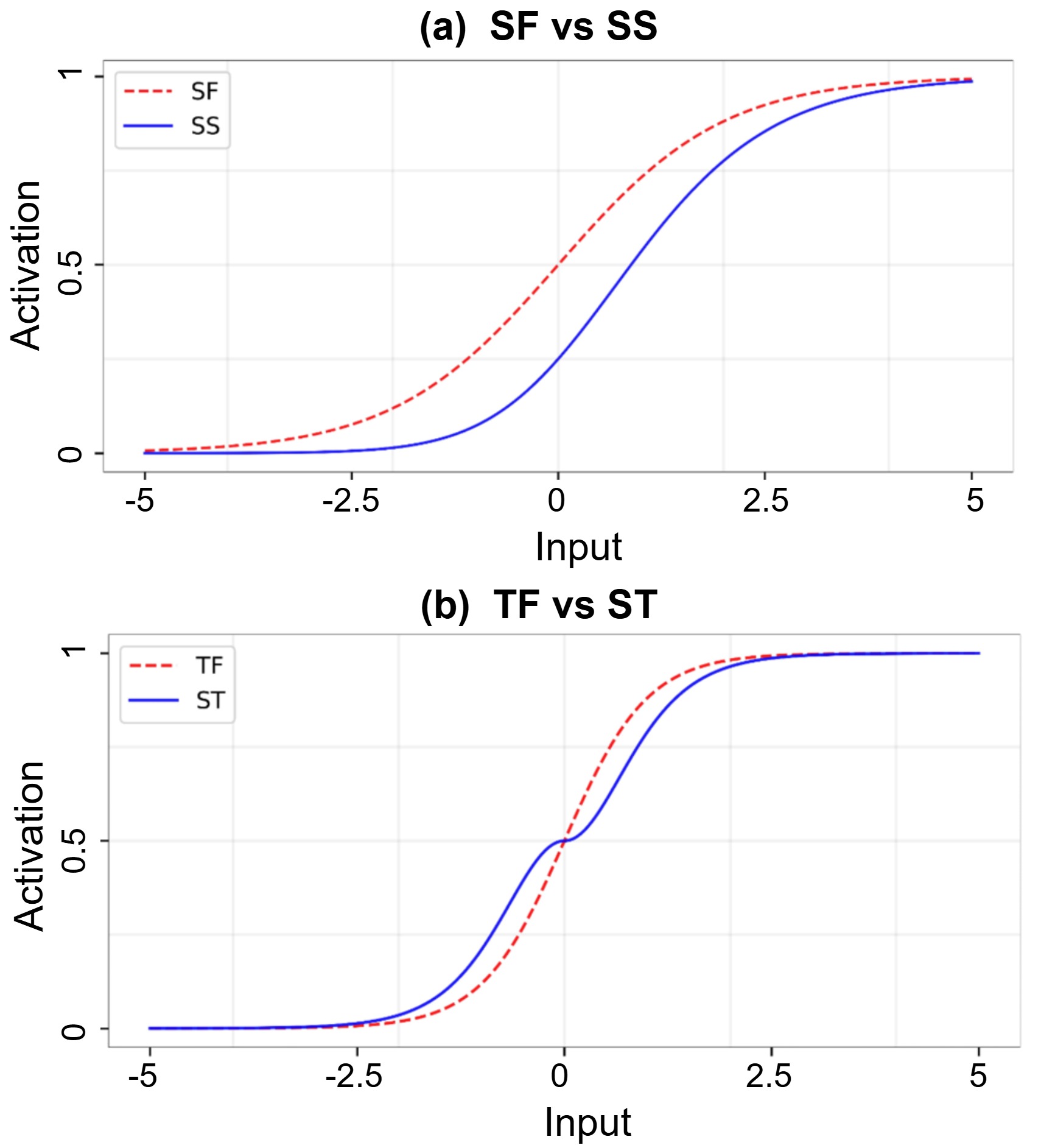}  
    \caption{Behaviour of (a) SS and (b) ST.}
    \label{fig:3}
 \end{figure}

The SS is achieved by squaring the generic sigmoid function (SF). It is defined by
\noindent
\begin{equation}
\label{eq:5} 
SS\left ( x \right ) = \left ( SF(x) \right )^{2}, \quad  where \quad SF(x) =\frac{1}{1+e^{-x}}.
\end{equation}
\noindent
The behaviour of SF and SS is displayed in Figure \ref{fig:3}a. The typical SF results in [0,1]. When SS is applied, the higher output probability value of SF is retained relatively higher when compared to the lower value getting very small. For example, $\left ( 0.9 \right )^{2}$ and $\left ( 0.1 \right )^{2}$ result in $0.81$ and $0.01$, respectively. This suggests that neurons are activated to focus on the larger value in retaining the sparse patterns. To emphasize the applicability of SS as activation function in sequential models, it is essential to discuss the properties of a generic activation function. They are boundedness, nonlinearity, continuity, and differentiability.    \\ [0.1cm] 
\textbf{(a) Range: }SS produces a similar S-shaped curve stretched between 0 and 1.\\ [0.1cm]
\textit{Proof: }To find the range of $SS(x)$, we show that $\forall_{x\in \mathbb{R}},0\leqslant SS(x)\leqslant 1$. Since SF is bounded by [0,1], squaring both sides results in [0,1] stretching the curve. \\ [0.1cm] 
\textbf{(b) Nonlinearity: }SS introduces more nonlinearity to the model than the traditional SF.\\ [0.1cm]
\textit{Proof: }Rate of change of SS is a function. When $\frac{d}{dx}SS(x)$ is taken, it results in $2*SF(x)*{SF}'(x)$, which is nonlinear, as shown in Figure \ref{fig:3}a. Hence, the SS has a more pronounced curvature in the middle of the curve compared to the SF. It means that it can map more complex sparse patterns in the data, as SS is more sensitive to small changes in the input.\\ [0.1cm] 
\textbf{(c) Continuity: }SS is continuous at all points in the curve. \\ [0.1cm]
\textit{Proof: }To prove $SS(x)$ is continuous, we consider the composition rule of continuity which states that if $g(x)$ is continuous at $x=a$ and $h(x)$ is continuous at $x=g(a)$, then the composite function $f(x)= g(h(x))$ is also continuous at $x=a$. In our case, we apply the composition rule to $g(x)=x^{2}$ and $h(x)= SF(x)$ to show that $SS(x)=g(h(x))$ is continuous. We know that $\forall_{x},h(x)$ and $g(x)$ are continuous. Therefore, by the composition rule for continuity, $\forall_{x},SS(x)= g(h(x))=(SF(x))^{2}$ is also continuous.  \\ [0.1cm] 
\textbf{(d) Differentiability: }SS is differentiable. \\ [0.1cm]
\textit{Proof: }To prove $SS(x)$ is differentiable $\forall x$, we show that there exists derivative at every point in its domain. Using the chain rule and the derivative of the logistic function, we find the derivative of SF and SS as follows,
\noindent
\begin{equation}
\label{eq:6} 
{SF}'(x) = SF(x)*(1-SF(x)).
\end{equation}
\noindent
\begin{equation}
\label{eq:7} 
{SS}'(x) = 2*SF(x)*{SF}'(x) .
\end{equation}
\noindent
Substituting (2) in (3), we get
\noindent
\begin{equation}
{SS}'(x) = 2*(SF(x))^{2}*(1-SF(x)).
\end{equation}
\noindent
Since SF is differentiable $\forall x$, the product of two differentiable functions is differentiable. Therefore, SS is also differentiable $\forall x$ in its domain.  
\subsection{Squared tanh (ST)}
ST is obtained by squaring the classical hyperbolic tangent function (TF) and mathematically defined by  
\noindent 
\begin{fleqn}
  \begin{equation}
    \begin{medsize}
    \label{eq:8} 
    \begin{split} 
    ST(x)=\begin{cases}
    -(TF(x))^2 & \text{ if } x < 0 \\ 
    (TF(x))^{2} & \text{ if } x \geqslant 0  
    \end{cases}, TF(x) = \frac{e^x-e^{-x}}{e^x+e^{-x}}.
    \end{split}
    \end{medsize}
    \end{equation}
\end{fleqn}
\noindent
Here, squaring the -ve value is returned as -ve of the squared value, as given in Eq. (5). Otherwise, the behaviour of ST is not adhered to the classical TF. The behaviour of TF and ST is shown in Figure \ref{fig:3}b, in which ST displays much higher nonlinearity away from the origin. As TF is modified, we discuss how it impacts and ensures the properties of activation functions.\\ [0.1cm] 
\textbf{(a) Range: }ST has a similar S-shaped curve like TF but with a narrow waist in the middle, as shown in Figure \ref{fig:3}b and it is bounded to [-1,1].\\ [0.1cm]
\textit{Proof: }To prove the boundedness of ST outcome [-1,1] $\forall x, x\in \mathbb{R}$, consider the following two cases. 
\begin{itemize}[noitemsep,topsep=0pt]
    \item For $x\geqslant 0$, since $TF(x)$ ranges [0,1], $(TF(x))^2$also ranges [0,1].   
    \item For $x<0$, since $TF(x)$ ranges [-1, 0), we use $-(TF(x))^2$which ranges [-1, 0).  
\end{itemize} 
Combining these cases together, Eq. (5) is bounded [-1,1] $\forall_x,x\in \mathbb{R}$.  \\ [0.1cm] 
\textbf{(b) Nonlinearity: }ST is non-linear.\\ [0.1cm]
\textit{Proof: }To prove ST is non-linear, we show that it does not satisfy the property of additivity. It means that $ST(a+b)\neq ST(a)+ST(b)$. For given $a,b\in \mathbb{R}$, nonlinearity is obvious for ST when $a,b \geqslant  0$. Therefore, we provide a clarity on nonlinearity for ST when $a,b<0$. From Eq. (5), we get $ST(a+b)=-(TF(a+b))^{2}$ when $a+b<0$. To prove nonlinearity, the following inequality holds $-(TF(a+b))^{2}\neq -(TF(a)+TF(b))^{2}$. Hence, additivity property is not applicable for the two cases considered above. \\[0.1cm]  
\textbf{(c) Continuity: }ST is continuous.\\ [0.1cm]
\textit{Proof: }To prove ST is continuous, it is sufficient to show that the ST is continuous at $x=0$. Consider the limit of $ST(x)$ as $x \rightarrow 0$ from the left and right side of a number line.
\noindent
\begin{itemize} 
    \item For $x \in \mathbb{R}^-$, \\ $\lim_{x \rightarrow 0}-(TF(x))^2=\lim_{x \rightarrow 0}-(0)^2=0$
    \item For $x \in \mathbb{R}^+$, \\ $\lim_{x \rightarrow 0}(TF(x))^2=\lim_{x \rightarrow 0}(0)^2=0$
\end{itemize}
\noindent
Since both the left and right limits are equal to 0, we can conclude that the ST is continuous at $x=0$. \\ [0.1cm] 
\textbf{(d) Differentiability: }ST is differentiable.\\ [0.1cm]
\textit{Proof: }To show that ST is differentiable at any $\mathbb{R}$, let us consider the chain rule, which states that if there is a composite function $f(g(x))$, then the derivative of $f(g(x))$ is given by $f'(g(x))*g'(x)$. Let $f(x)= TF(x)$ and $h(x)=2I(x)-1$, where $I(x)$ is the indicator function. Then the ST can be expressed as $g(x) =f(x)^2*h(x)$ and the derivative of $g(x)$ is given by 
\noindent
\begin{equation}
\label{eq:11}
g'(x) =2*f(x)*f'(x)*h(x)+f(x)^2*h'(x).
\end{equation}
\noindent
Using the fact that $f'(x)=Sech^2(x)$, $h(x)$ is a piecewise constant and $h'(x)$ is 0 almost everywhere (except at $x=0$), we simplify Eq. (6) as 
\noindent
\begin{equation} 
g'(x)=2TF(x)*Sech^2(x)[2I(x)-1], \quad if \quad x\neq 0.
\end{equation}
\noindent
It turns out that $g(0)=0$. Note that $Sech^2(x)$ and $2I(x)-1$ are both continuous functions at any $\mathbb{R}$, so their product is also continuous at any $\mathbb{R}$. Thus ST is differentiable at any $\mathbb{R}$, except possibly at $x=0$, where we have a corner point. 

\subsection{SST (SS+ST)}
To improve the performance of the classical GRU further, the SST is conceived based on observing and understanding their role in the gating mechanism, as well as their properties discussed in the previous sections. The basic concept of the SST is that to make the small values even smaller and the large values even larger, by squaring the output of SF in $r_{t}$ and $z_{t}$, and TF in the dense layers, such that SST
\begin{itemize}[noitemsep,topsep=0pt]
    \item improve gradient sensitivity and learning dynamics in sparse regimes.
    \item improves selective retention of informative signals and helps better recall information from previous $t$. 
    \item provides a greater degree of nonlinearity, which is particularly important for $r_{t}$ to capture complex patterns and determine what information should be retained.
    \item emphasizes the robustness of GRU cells to noise input values by focusing on important information, particularly for $z_{t}$, which determines how much of $h_{t-1}$ to retain.  
\end{itemize}
It is important to observe that we did not replace the default TF in $\widetilde{h}_{t}$ of the GRU, as it can make it difficult to update $\widetilde{h}_{t}$ over time. Also, the reason for not using the widely used ReLU activation function in the dense layer is to avoid the \enquote{zero-gradient} problem for the negative inputs and to provide smooth gradient transitions to improve the overall stability and convergence rate of the neural network. Even though ReLU has been shown to work well in many cases, the use of ST in the dense layers was superior to ReLU in terms of test accuracy. Ultimately, as expected, SST improved the overall performance and learning efficiency of the neural networks.  

\section{Experiments} 
We perform controlled comparisons between standard GRU and GRU-SST to isolate the effect of contrast-enhanced gating under low-data conditions. Models use identical architectures and training settings to ensure fair comparison. We comprehensively analyse the effectiveness of SST compared to traditional functions (sigmoid and tanh) on different tasks, such as sign language recognition, gait classification, and human activity recognition tasks, all of which involve temporal modelling of human motion signals commonly studied in computer vision and multimodal perception literature. 

\subsection{Sign Language Recognition }
Sign language recognition is a critical task in human-computer interaction, enabling seamless communication with individuals who are deaf. For the experiments, the configuration is set according to \cite{ref14}. The authors employ mediapipe optimized gated recurrent unit (MOPGRU) model to improve the performance with limited datasets. It consisted of real-time video recordings of 13 distinct Indian sign gestures as mentioned in \cite{ref14}. Each sign gesture was represented by a collection of 30 videos, each comprising 30 frames, and all videos shared a uniform size of 640 × 480 pixels. To understand the data better, PCA visualization tool was adopted. From the plot, we observed that there exists sparse patterns in the ISL data, as shown in Figure \ref{fig:1}. To address data sparsity and improve predictive performance, we incorporated the SST activation function into the MOPGRU model, resulting in the MOPGRU-SST configuration. The performance of MOPGRU-SST is compared against the standalone MOPGRU model, and the results are discussed in the following section.   \\[0.2cm]
\textbf{T-SNE visualization of learned representations:} The visualization of hidden layer outputs as shown in Figure \ref{fig:4a} and \ref{fig:4b} offers a compelling view into the semantic representation capabilities of the MOPGRU and MOPGRU-SST models. 
\begin{figure}[t!]
\centering
    \begin{minipage}{0.5\textwidth}
        \centering
        \includegraphics[width=0.93\textwidth]{Fig5a.jpg}
        \subcaption[first caption.]{MOPGRU}
        \label{fig:4a}
    \end{minipage} \\
    \begin{minipage}{0.51\textwidth}
        \centering
        \includegraphics[width=0.93\textwidth]{Fig5b.jpg}
        \subcaption[second caption.]{MOPGRU-SST}
        \label{fig:4b}
    \end{minipage}%
    \vspace{-5pt}
\caption{T-SNE visualization of hidden layer outputs.} 
\label{fig:4}
\vspace{-5pt}
\end{figure}
\begin{figure}[t!]
\centering
    \begin{minipage}{0.5\textwidth}
        \centering
        \includegraphics[width=0.93\textwidth]{Fig6a.jpg}
        \subcaption[first caption.]{MOPGRU}
        \label{fig:5a}
    \end{minipage} \\
    \begin{minipage}{0.51\textwidth}
        \centering
        \includegraphics[width=0.93\textwidth]{Fig6b.jpg}
        \subcaption[second caption.]{MOPGRU-SST}
        \label{fig:5b}
    \end{minipage}%
    \vspace{-5pt}
\caption{T-SNE visualization of dense layer outputs.} 
\label{fig:5}
\vspace{-10pt}
\end{figure}
In the t-SNE plots, we observe that MOPGRU forms broader clusters for words such as \enquote{think}, \enquote{phone call}, and \enquote{meet}, which could be indicative of its capacity to capture and represent semantic relationships. In contrast the separation is more prominent in MOPGRU-SST, signifying its ability to encode knowledge representations in a highly distinguished manner. This differentiation can enhance recognition capabilities by efficiently mapping varying inputs to more granular model concepts. Analysing the dense layer visuals reveals similar insights, with MOPGRU-SST in Figure \ref{fig:5b} producing clearer delineation between clusters representing different terms. The evident distinction of semantic groupings compared to MOPGRU in Figure \ref{fig:5a} indicates that MOPGRU-SST develops richer hierarchical data representations. Overall, by comparing the test accuracy of both models, MOPGRU-SST achieved a higher accuracy of 100\% than baseline MOPGRU which achieved 95\%. Although near-perfect accuracy is observed on this dataset, the number of gesture samples per class is limited and the task may be relatively structured in feature space. Therefore, this result should be interpreted as evidence of improved separability on a controlled low-data benchmark rather than as a definitive indication of general superiority. Evaluating SST on larger and more diverse gesture datasets remains an important direction for future work. 

\subsection{Gait classification}
We utilize an accelerometer-based gait dataset from \cite{ref24} comprising raw acceleration sequences collected from 244 subjects during a 3-minute walking task. The dataset provides a robust test application for evaluating recurrent models for classifying dynamic human activities that are sequential in nature. The original dataset, although suitable for analysing age-related differences in gait patterns as explored in \cite{ref24}, does not contain sufficient sparsity to assess model robustness under missing data conditions within data constraints. Since real-world sensor feeds often encounter missing information, we introduce sparsity to simulate this practical challenge. Specifically, 20\% of values in the 1024-sample input sequences were randomly zeroed out to induce missing data points across temporal dimensions. In the induced sparse dataset, a sharp peak at zero indicates significant conversion of values to a single number, effectively creating sparse sequences. With 1460 sequences originally, this process generates 1168 non-sparse and 292 sparse patterns in the final training dataset. The incorporation of sparse data simulations allows comprehensive benchmarking of SST-powered GRU models against traditional GRU in handling sparse temporal inputs. Training the model, the experimental setup and model training parameters follow the same as mentioned in \cite{ref24} with the additional incorporation of SST activation function in the classical GRU. For evaluating the classification performance, widely used evaluation metrics such as accuracy, precision, recall and F1 score were used. In addition, receiver operating characteristic (ROC) curves and the area under the curve (AUC) were generated. \\ [0.1cm]
\textbf{Classification performance:} The integration of the proposed SST activation consistently improves GRU performance over the baseline across evaluation metrics as observed in Table \ref{tab:1}. GRU-SST achieves a superior test accuracy of 84.3\% compared to classical GRU's 79.8\%. Precision increases from 78.9\% to 87.7\% with SST incorporation. Further, GRU-SST obtains an AUC of 0.92, substantially superior to the baseline GRU's 0.86 AUC, indicating higher discrimination between classes. With consistent gains in accuracy, precision, recall, and AUC, SST demonstrates concrete effectiveness in addressing GRU limitations when learning from sparse temporal data. The results validate SST's strengths in transforming classical recurrent models within constraints to boost predictive performance. \\ [0.1cm]
\begin{table}[t!]
\centering
\scriptsize
\caption{Classification performance of GRU and GRU-SST.}
\vspace{-5pt}
\label{tab:1} 
\begin{tabular}{|l|c|c|c|c|c|}
\hline
\multicolumn{1}{|c|}{\textbf{Model}} & \textbf{Test Accuracy} & \textbf{Precision} & \textbf{Recall} & \textbf{F1-score} & \textbf{AUC} \\ \hline
GRU                                  & 79.8\%                 & 78.9\%             & 77.8\%          & 78.3\%            & 0.86         \\ \hline
GRU-SST                              & 84.3\%                 & 87.7\%             & 77.4\%          & 82.2\%            & 0.92         \\ \hline
\end{tabular} 
\vspace{-15pt}
\end{table}
\begin{figure}[b!]
\vspace{-10pt}
\centering
    \begin{minipage}{0.53\textwidth}
        \centering
        \includegraphics[width=0.65\textwidth]{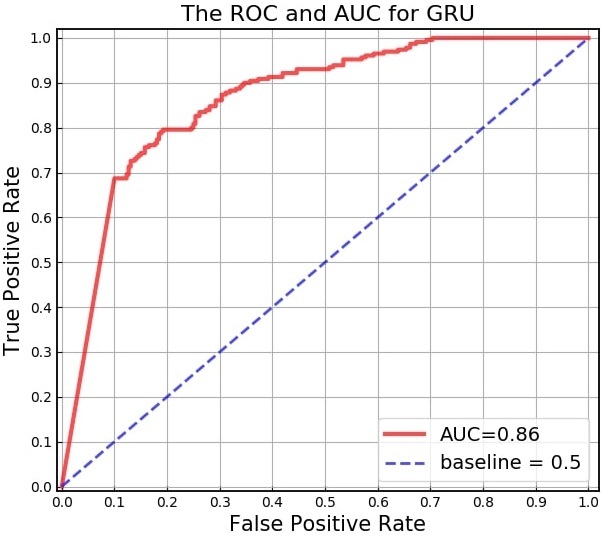}
        \subcaption[first caption.]{GRU}
        \label{fig:6a}
    \end{minipage} \\
    \begin{minipage}{0.53\textwidth}
        \centering
        \includegraphics[width=0.65\textwidth]{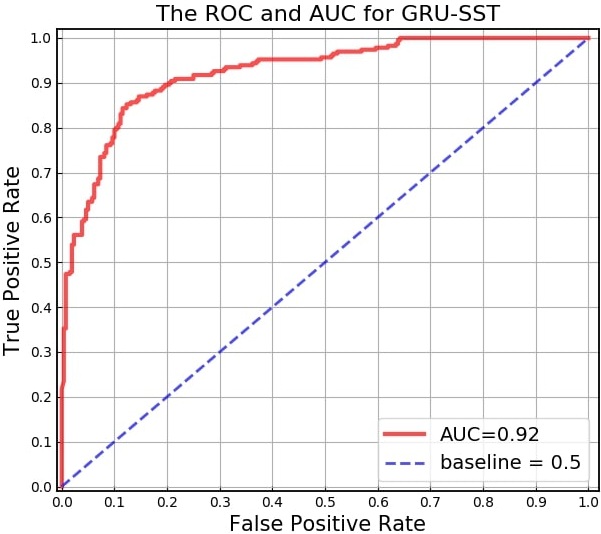}
        \subcaption[second caption.]{GRU-SST}
        \label{fig:6b}
    \end{minipage}%
    \vspace{-5pt}
\caption{ROC curves.} 
\label{fig:6}
\end{figure}
\noindent
\textbf{ROC and AUC curve analysis:} The ROC curve shown in Figure \ref{fig:6} illustrates the performance of baseline GRU and GRU-SST models on a classification task. The red line represents the true positive rate (TPR) versus the false positive rate (FPR) at various threshold levels. An ideal model would have a curve that reaches towards the top left corner, indicating a high TPR and low FPR. The ROC analysis provides evidence on SST's efficacy with GRU-SST producing a higher AUC of 0.92 over the baseline's 0.86. This increased AUC value from 0.86 to 0.92 signifies a notable improvement in model performance, highlighting the potential of SST to improve the robustness and predictive accuracy of GRU networks in classification performance as illustrated by the ROC curve.

\subsection{Human activity recognition}
Human activity recognition (HAR) using wearable sensors provides an attractive test application for evaluating sequential modelling techniques. Sensor feeds capturing motions over time exhibit temporal dependencies well-suited to GRU networks. We leverage two standard HAR datasets to evaluate our proposed SST activation in enhancing GRU models. The first is the WISDM dataset \cite{ref30} comprising 1,098,207 samples from 36 subjects performing daily activities like walking, jogging, sitting, and standing. The second is the UCI Smartphone dataset \cite{ref31} containing 10,299 instances of sensor readings from 30 volunteers aged 19 to 48 years executing 6 standard activities - standing, sitting, lying, walking, upstairs and downstairs using a mobile phone. Each subject performed the protocol twice with varying device placements. 50Hz triaxial acceleration and angular velocity data were recorded and manually labelled. We induced 20\% sparsity into both datasets to simulate real-world sensor feed challenges. For training, we followed \cite{ref32} and compared classical GRU and GRU-SST performance for both datasets. The comparative evaluation results summarized in Table \ref{tab:2} clearly validate the effectiveness of the proposed SST activation in enhancing classical GRU models for human activity recognition, consistently across both datasets. \par

\begin{table}[H]
\vspace{-5pt}
\centering
\scriptsize
\caption{Comparative performance of GRU and GRU-SST models on WISDM and UCI-HAR datasets.}
\label{tab:2} 
\begin{tabular}{|l|c|c|c|}
\hline
\multicolumn{1}{|c|}{\textbf{Model}} & \textbf{Dataset} & \textbf{\begin{tabular}[c]{@{}c@{}}Training \\ Accuracy (\%)\end{tabular}} & \textbf{\begin{tabular}[c]{@{}c@{}}Testing \\ Accuracy (\%)\end{tabular}} \\ \hline
GRU                                  & WISDM            & 99.5                                                                       & 97.08                                                                     \\ \hline
GRU-SST                              & WISDM            & 100                                                                        & 99.28                                                                     \\ \hline
GRU                                  & UCI-HAR          & 99                                                                         & 93.08                                                                     \\ \hline
GRU-SST                              & UCI-HAR          & 100                                                                        & 98.30                                                                     \\ \hline
\end{tabular}
\vspace{-5pt}
\end{table}

On sparse WISDM accelerometer data, SST allows GRU models to achieve close to 100\% training accuracy. More importantly, test accuracy increases over 2\% with GRU-SST, reaching 99.28\% despite introduced sparsity. Similarly on Smartphone data, SST again shows generalization capability improving GRU test accuracy by 5\% to 98.3\%. The results validate SST's effectiveness in tackling GRU limitations on multivariate time-series across application domains. By reshaping classical activations, SST unlocks substantial accuracy and convergence gains. The 100\% training accuracy observed in some HAR settings should be interpreted with caution, as these benchmarks are relatively well studied and may permit strong fitting under the current experimental setup. More informative than the training accuracy is the consistent improvement in test performance, which suggests that SST improves gating behaviour under the evaluated conditions. Nevertheless, broader validation across additional datasets and repeated trials would further strengthen conclusions about robustness and generalization. 

\subsection{Time-series forecasting}
In financial forecasting, the ability to accurately predict future prices of commodities like gold is a testament to a model's predictive prowess. Gold, a commodity of high economic interest, presents a challenging yet valuable dataset for time-series analysis. The historical gold price dataset from Yahoo Finance, with its comprehensive range of data points from 2000 to 2026, was chosen for its complexity and relevance. The dataset's granular details, including open, high, low, close, and adjusted close prices, offer a multifaceted view of market behaviour, allowing the model to learn from sparse temporal patterns. The choice to use this dataset aligns with our objective to assess SST's ability to enhance GRU model performance. The daily fluctuations and trends encapsulated within gold price data require a model that discerns long-term dependencies. SST's design to amplify gradient flow and improve information retention is put to the test, making this dataset suitable for demonstrating SST's potential to refine predictions in a complex and unpredictable domain like financial time-series. For training, we utilized the model configuration from \cite{ref33} based on GRU architecture by replacing the LSTM layers. To analyze SST improvements, mean squared error (MSE) loss was calculated between predicted and ground truth gold prices over the test period. As evidenced in Table \ref{tab:3}, SST integration leads to substantial reductions in error metrics consistently across training and validation. GRU-SST lowers MSE compared to baseline GRU in terms of training and validation loss by over 20\%. The test set MSE declines even further from 0.28 to 0.08. These consistent MSE reductions across phases validate SST's capability in accurately modelling temporal dependencies for financial forecasting with GRU. By tuning activations to sequential data, SST unlocks substantial predictive performance gains over classical approaches.
\vspace{-5pt}
\begin{table}[t!]
\centering
\scriptsize
\caption{Loss and MSE comparison for GRU and GRU-SST on gold price forecasting.}
\label{tab:3} 
\begin{tabular}{|l|c|c|c|}
\hline
\multicolumn{1}{|c|}{\textbf{Model}} & \textbf{Training Loss} & \textbf{Validation Loss} & \textbf{Test Loss} \\ \hline
GRU                                  & 0.70                   & 0.08                     & 0.28         \\ \hline
GRU-SST                              & 0.54                   & 0.06                     & 0.08         \\ \hline
\end{tabular}
\vspace{-5pt}
\end{table}

\section{Conclusion} 
We presented squared sigmoid-tanh (SST), a zero-parameter modification to GRU gating that increases activation contrast by squaring gate outputs. SST is designed for low-data and sparse-signal settings where standard sigmoid/tanh gating may be overly smooth. Across diverse sequential tasks, SST-GRU consistently improved performance over a vanilla GRU without altering the GRU architecture. On sign language recognition, SST achieved 100\% test accuracy compared to the baseline on this dataset. On gait classification, SST improved test accuracy from 79.8\% to 84.3\%. On human activity recognition, SST increased test accuracy to 99.28\% (gain $>$2.2\%) and reached 98.3\% on the Smartphone dataset (gain 5\%). For forecasting, SST reduced gold price prediction MSE from 0.28 to 0.08, accompanied by lower training and validation losses. These results show that a targeted nonlinearity change in GRU gating can improve performance in data-scarce settings with negligible computational cost. \\[0.2cm] 
\textbf{Limitations}. Our evaluation focuses on GRU-based pipelines and low-data benchmarks; future work should test SST under broader distribution shifts, longer sequence horizons, and stronger baselines such as lightweight Transformers or state-space models. Our study isolates the effect of contrast-enhanced gating within GRU architectures under limited-data conditions. Broader comparisons with alternative recurrent models (e.g., LSTM), activation variants (e.g., hard-sigmoid or parameterized nonlinearities), and modern sequence models such as Transformers or state-space approaches remain future work. Because SST sharpens gate responses, very small activations may be further suppressed, which could affect extremely long or weak-signal sequences.   





\end{document}